\pdfoutput=1  
\RequirePackage{fix-cm}
\documentclass{article} 
\usepackage{iclr2027_conference,times}

\usepackage{amsmath,amssymb,amsthm,mathtools}
\usepackage[T1]{fontenc}
\usepackage{newtxtext,newtxmath}
\usepackage{bm}

\usepackage{amsmath,amsfonts,bm}

\def\eqref#1{equation~\ref{#1}}

\def\1{\bm{1}}

\DeclareMathAlphabet{\mathsfit}{\encodingdefault}{\sfdefault}{m}{sl}
\SetMathAlphabet{\mathsfit}{bold}{\encodingdefault}{\sfdefault}{bx}{n}

\usepackage{xcolor}
\usepackage{graphicx}
\usepackage{caption}
\usepackage{subcaption}
\usepackage{placeins}
\usepackage{wrapfig}
\usepackage{booktabs}
\usepackage{multirow}
\usepackage{tabularx}
\usepackage{makecell}
\usepackage{array}
\usepackage{colortbl}
\usepackage{pifont}
\usepackage[most]{tcolorbox}

\usepackage{hyperref}
\usepackage{url}

\definecolor{YesGreen}{HTML}{238653}
\definecolor{NoRed}{HTML}{C34A4A}
\definecolor{ourscolor}{HTML}{F8E9EA}

\newcommand{\yes}{\textcolor{YesGreen}{\ensuremath{\bigcirc}}}
\newcommand{\no}{\textcolor{NoRed}{\ensuremath{\times}}}

\newcommand{\systems}{\textsc{SALT}}
\hypersetup{hidelinks}
\title{Adapting Vision-Language-Action Models\\to Unknown Visual Disruptions During Execution}

\author{%
  Ahin Lee \quad Jinwoo Seo \quad Youngsoo Jang \quad Taesik Gong \\
  UNIST, Ulsan, Republic of Korea \\
  \texttt{\{ahin, jinwoo, youngsoo.jang, taesik.gong\}@unist.ac.kr} 
}

\iclrfinalcopy 

\begin{document}

\maketitle
\lhead{Preprint.}

\begin{abstract}
Visual disruptions can arise while a robot is executing a task, leaving a
vision--language--action (VLA) policy to respond without knowing the
disruption type or timing.
We introduce \textbf{Self-supervised Adaptation from Leftover Trajectories (SALT)}, which uses the leftover trajectory, the unexecuted part of the previous
action chunk, as self-supervision for test-time adaptation.
Because consecutive chunks overlap in time, the leftover provides a temporally
aligned target for the current prediction over the same future control interval.
At the onset of a visual shift, the leftover can retain a plan formed before the
corruption, so updating the policy toward it anchors the adaptation across the shift (Transition Anchoring).
SALT keeps the adapted policy and regenerates the current chunk, whose leftover
becomes the target at the next replan, carrying the correction forward along the
execution trajectory (Sequential Correction Propagation).
Supervision comes entirely from the policy's own predictions, requiring no disruption annotations, expert actions, or target-domain demonstrations, and a lightweight adaptation gate calibrated only on nominal trajectories decides when updates begin.
On LIBERO-10, SALT increases average success across five persistent visual
corruptions from 43.9\% to \textbf{53.2\%} with SmolVLA and from 58.7\% to
\textbf{66.0\%} with GR00T N1.7, while largely preserving nominal performance.
On a real robot, it raises task progress averaged over digital and physical
disruptions from 0.49 to \textbf{0.61}.
\end{abstract}
\section{Introduction}
\label{sec:introduction}

\begin{figure*}[t]
    \centering
    \includegraphics[width=\textwidth]{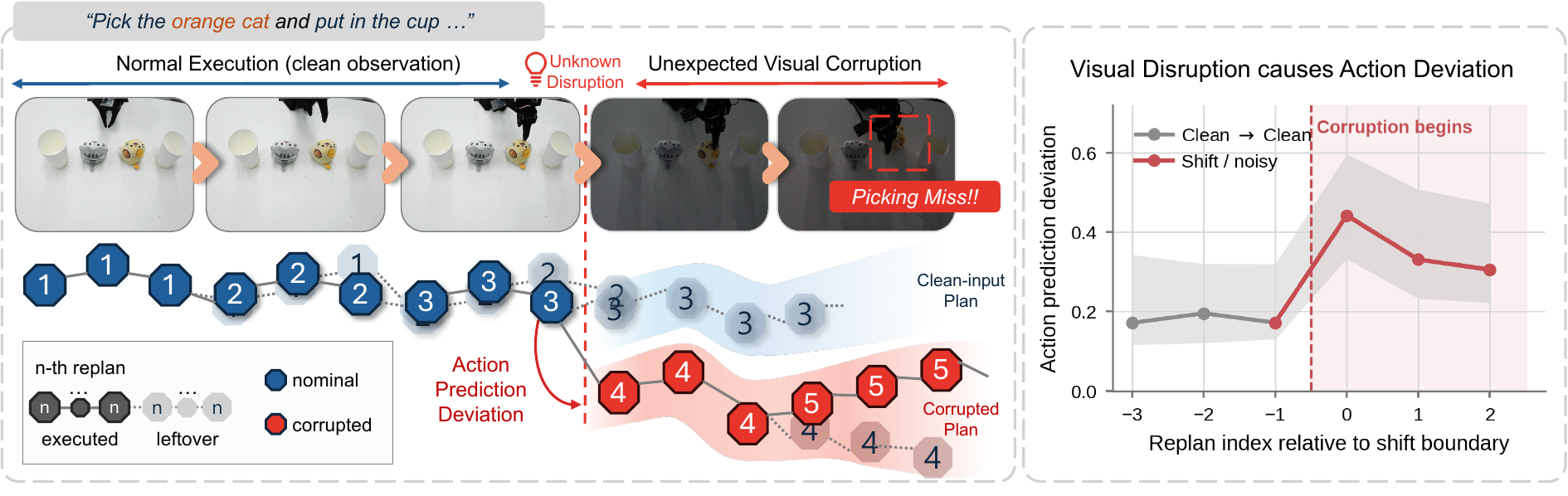}
    \caption{\textbf{Visual disruption during an ongoing task.}
    A VLA begins executing under nominal observations, but an unexpected visual corruption can arise during execution and alter its subsequent action predictions. \textbf{Left:} an example of task failure under the visual disruption. \textbf{Right:} deviation between each new action chunk and the aligned leftover of the previous chunk. It stays small under nominal observations (gray) but jumps at the first corrupted replan (red). See Appendix~\ref{app:fig1_diagnostic} for details.
    }
    \label{fig:lava_problem}
\end{figure*}

Vision--Language--Action (VLA) models map visual observations and natural-language instructions to robot actions, enabling a single policy to perform diverse manipulation tasks~\citep{brohan2023rt2,kim2024openvla,ghosh2024octo,black2024pi0,bjorck2025groot}. Despite these capabilities, reliable deployment remains challenging when unexpected changes in visual conditions occur during task execution~\citep{fei2026liberoplus, morgan2026colosseum}. 
Changes in lighting, image quality, or camera visibility can perturb the observation stream on which the policy relies and abruptly change its action predictions (Figure~\ref{fig:lava_problem}, right), which can substantially degrade task performance.
We refer to such execution-time events as \emph{visual disruptions}, which induce a \emph{visual shift} from the nominal observation condition while the task is already underway.

Test-time adaptation (TTA) provides a framework for adapting pretrained models to distribution shifts using signals available at deployment, such as predictive entropy~\citep{wang2021tent,gong2023sotta} or teacher predictions~\citep{wang2022cotta,dobler2023rmt}. 
Existing VLA adaptation methods rely on signals prepared outside the current episode, such as delayed feedback, learned progress estimates, or target-domain demonstrations~\citep{zang2026pdf,kang2026dart,li2026fla,bai2025evolve}.
When a disruption arises during execution with unknown type and timing (Figure~\ref{fig:lava_problem}, left), such resources are not available, and the policy must respond within the ongoing rollout using only what execution has already produced.
We ask: \textbf{How can a VLA adapt online to an unknown visual disruption using only information already produced during execution?}

Our key observation is that action-chunked execution already exposes a useful learning signal within the rollout.
At each replan, a VLA predicts an action chunk~\citep{zhao2023act} but executes only a prefix~\citep{black2024pi0,liu2025bidirectional}. We refer to the remaining, unexecuted action sequence from the previous chunk as a \emph{leftover trajectory}. 
Because the leftover and the prefix predicted at the next replan cover the same future control interval, the policy obtains temporally aligned supervision without any new expert label. At the shift boundary, this leftover was planned from a nominal observation, so it retains the intended plan that the corrupted prediction has lost.

Building on this observation, we introduce 
\textbf{S}elf-supervised \textbf{A}daptation from \textbf{L}eftover \textbf{T}rajectories (\textbf{SALT}), a leftover-supervised mechanism for test-time adaptation of VLAs within the ongoing episode. Once adaptation is activated, SALT uses the previous chunk's leftover as self-supervision under the current policy input, updates the policy, and regenerates the action chunk before execution. The same supervision plays two complementary roles. Across the visual shift, the nominal-side leftover can provide \textbf{Transition Anchoring}. Thereafter, SALT retains the adapted policy state and refreshes the leftover target at successive replans, yielding \textbf{Sequential Correction Propagation} along the execution trajectory. To deploy SALT when disruption timing is unknown, we use a lightweight adaptation gate calibrated only on nominal trajectories. The gate determines when these leftover-supervised updates begin without requiring annotations of the disruption type or timing.

We evaluate \systems{} on LIBERO-10 tasks~\citep{liu2023libero} under five persistent visual corruptions from LIBERO-Plus~\citep{fei2026liberoplus}. With SmolVLA~\citep{shukor2025smolvla}, \systems{} raises average success across corrupted conditions from 43.9\% to \textbf{53.2\%}, and with GR00T N1.7~\citep{bjorck2025groot}, it raises success from 58.7\% to \textbf{66.0\%}, while largely preserving nominal performance. 
The leftover target approaches an oracle target that is unavailable at
deployment. On SmolVLA, \systems{} reaches 53.2\%, within 3.7 points of
supervising with the policy's clean-input prediction (56.9\%), whereas other
nominal targets, such as another episode's plan, remain at or below the frozen
level (43.9\%). On a real robot, \systems{} raises task progress from 0.46 to \textbf{0.61} under digital disruptions and from 0.51 to \textbf{0.62} under physical disruptions such as switching off the lights or covering a camera. 
Because \systems{} relies only on the temporal overlap between consecutive action chunks for supervision, its core mechanism is compatible with action-chunked backbones without requiring architectural modifications.

\section{Related Work}
\label{sec:related}

\paragraph{Robustness of VLA models under visual shifts.}
VLA policies have shown strong performance across diverse manipulation
tasks~\citep{brohan2023rt2,kim2024openvla,black2024pi0,bjorck2025groot},
but visual shifts such as sensor noise and lighting changes can sharply
degrade execution~\citep{fei2026liberoplus,morgan2026colosseum}. Prior work
improves visual robustness through training on perturbed images~\citep{xie2026strongvla}
or learned visual adapters~\citep{fu2026stablevla}, intervenes on or restores
observations at inference time while keeping the policy fixed~\citep{hancock2024byovla,
guarnizo2026crt}, or adapts from target-domain demonstrations~\citep{kang2026dart,
li2026fla}. Unlike these approaches, SALT updates the policy when a visual
shift of unknown type and timing arises mid-episode, using only observations
and action predictions from the ongoing rollout without target-domain
demonstrations or corruption-specific training.

\paragraph{Test-time adaptation of policies.}
Test-time adaptation methods for visual perception update models from
unlabeled test data through entropy-based objectives~\citep{wang2021tent,
gong2023sotta}, teacher predictions~\citep{wang2022cotta,dobler2023rmt},
or activation-statistics matching~\citep{mirza2023actmad}. These objectives
do not directly provide targets for temporally aligned continuous action
chunks. VLA-specific approaches instead use delayed episode feedback~\citep{zang2026pdf},
learned progress estimates~\citep{bai2025evolve,liu2026ttvla}, predicted and
observed future images in visual-foresight policies~\citep{park2026t3vf},
a proxy task trained before deployment~\citep{zhang2026tttvla}, or additional
exploratory rollouts~\citep{chen2026t2vla}. 
Several of these assume discrete action tokens or a future-image head and do
not transfer directly to recent flow-based action experts. Unlike methods that
depend on such feedback, auxiliary components, or extra rollouts, SALT uses
temporally aligned leftover actions already generated during the same
episode as targets for online updates, and applies to any action-chunked
backbone.

\paragraph{Reusing overlapping action chunks.}
Prior work uses overlapping action chunks to improve execution: ACT
temporally ensembles predictions~\citep{zhao2023act}, SEAM smooths chunk
transitions~\citep{zhan2026seam}, BID selects chunks consistent with previous
actions~\citep{liu2025bidirectional}, and RTC freezes actions committed during
inference latency~\citep{black2025rtc}. AAC adjusts the execution
horizon from action uncertainty~\citep{liang2026adaptive}. 
These methods reuse overlapping predictions to improve inference-time continuity, chunk selection, or execution while keeping the policy parameters fixed. In contrast, \systems{} uses temporally aligned leftover actions as self-supervision for online policy updates under within-episode visual shifts, which to our knowledge has not been explored before.

\section{Preliminaries and Problem Setting}
\label{sec:prelim}

\subsection{Action-Chunked VLA Execution}
\label{sec:prelim_action_chunking}

Let $r_t$ denote the $t$-th replan. At replan $r_t$, a VLA policy $\pi$ predicts an action chunk from visual observation $o_t$, proprioceptive state $s_t$, and task instruction $g$:
\begin{equation}
    \mathbf{A}_t
    = [\mathbf{a}_{t,0},\ldots,\mathbf{a}_{t,H-1}]
    \sim \pi(\cdot \mid o_t,s_t,g),
    \qquad
    \mathbf{A}_t \in \mathbb{R}^{H\times d},
\end{equation}

\begin{wrapfigure}{r}{0.40\linewidth}
    \vspace{-1.0em}
    \centering
    \includegraphics[width=\linewidth]{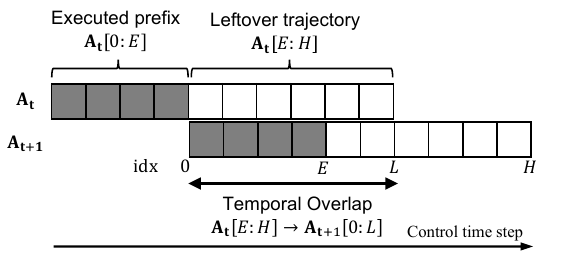}
    \captionsetup{font=small}
    \caption{\textbf{Temporal overlap between consecutive chunks.} The leftover trajectory $\mathbf{A}_t[E{:}H]$ and the next chunk's prefix $\mathbf{A}_{t+1}[0{:}L]$ cover the same future control interval. Indices mark action positions in $\mathbf{A}_{t+1}$, so position $j$ of $\mathbf{A}_{t+1}$ and position $E+j$ of $\mathbf{A}_t$ refer to the same future control step.}
    \label{fig:action_chunk_overlap}
    \vspace{-3em}
\end{wrapfigure}
where $H$ is the prediction horizon and $d$ is the action dimension. Executing only the first $E<H$ actions before replanning can help the robot respond to new observations~\citep{liu2025bidirectional,liang2026adaptive}. With $L=H-E$, the unexecuted \emph{leftover trajectory} $\mathbf{A}_t[E{:}H]$ and the next chunk's prefix $\mathbf{A}_{t+1}[0{:}L]$ cover the same future control interval (Figure~\ref{fig:action_chunk_overlap}).
This temporal overlap provides the basis for the online self-supervision introduced in Section~\ref{sec:method}.

\subsection{Visual Disruptions During Execution}
\label{sec:prelim_visual_shift}

We consider episodes that begin under nominal visual conditions. An unknown
\emph{visual disruption} may change the visual input during execution.
We call the resulting deviation a \emph{visual shift}. Let $r_{t^*}$ denote the first replan affected by this shift,
marking the \emph{shift boundary} from nominal visual observations
$o_t^{\mathrm{nom}}$ to shifted visual observations
$o_t^{\mathrm{shift}}$. The task instruction, proprioception, embodiment, and action space
remain unchanged. The policy's visual input is
\begin{align}
    o_t =
    \begin{cases}
        o_t^{\mathrm{nom}}, & t < t^*,\\
        o_t^{\mathrm{shift}}, & t \ge t^*.
    \end{cases}
    \label{eq:visual_shift}
\end{align}
At deployment, neither the disruption type nor the shift boundary $t^*$
is known. The policy can use its current inputs and execution history,
but has no paired nominal observation, expert action, or prior
shifted-domain data. We allow access to a small set of nominal trajectories before deployment. The goal is to preserve the policy's nominal task
behavior under the shifted visual input without access to the corresponding
nominal observation.

\section{Method}
\label{sec:method}

\begin{figure}[t]
    \centering
    \includegraphics[width=\linewidth]{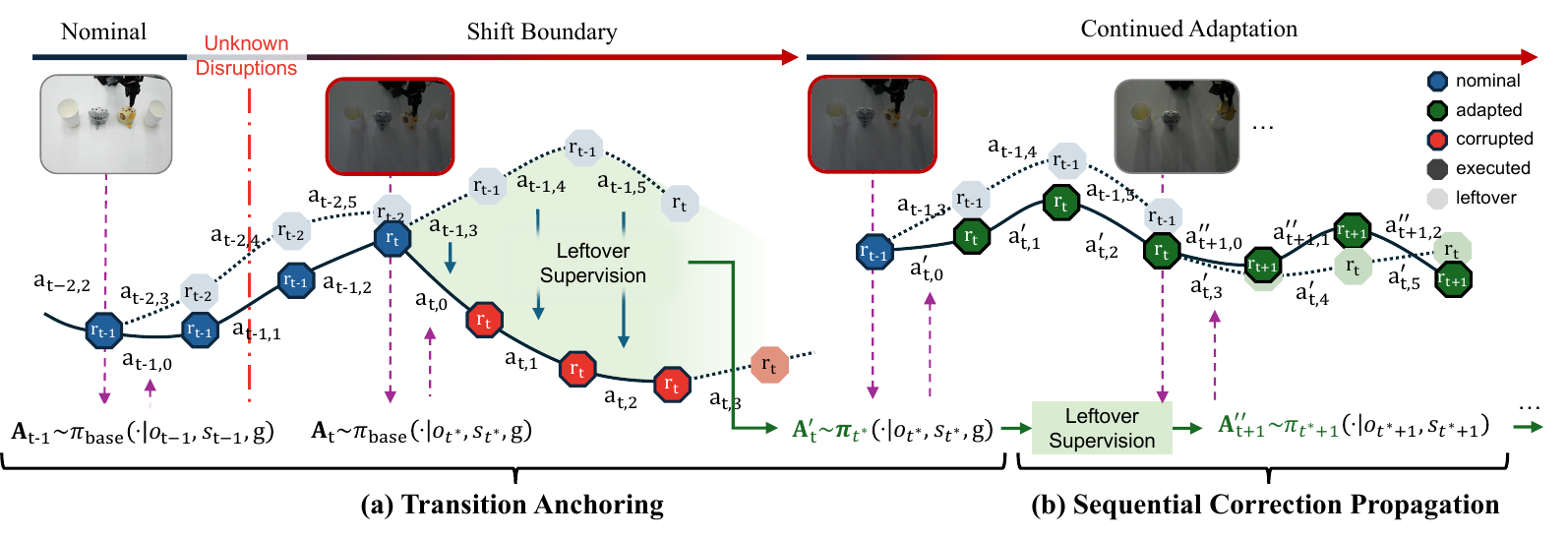}
    \caption{\textbf{SALT adapts to a visual disruption during execution.} Illustrated with a chunk of $H=6$ actions, of which $E=3$ are executed and $L=3$ remain as the leftover. (a) \textbf{Transition Anchoring}: at the shift boundary, the leftover of the chunk planned from the last nominal observation serves as the supervision target. SALT updates the policy under the shifted observation toward this leftover, pulling it toward the plan formed before the shift. (b) \textbf{Sequential Correction Propagation}: at later replans, the target is the leftover of the chunk regenerated by the adapted policy at the previous replan. Because the adapted policy is carried forward, each update refines the correction along the execution trajectory.}
    \label{fig:method_overview}
\end{figure}

We propose \textbf{Self-supervised Adaptation from Leftover Trajectories (SALT)},
 which repurposes the leftover trajectory introduced in Section~\ref{sec:prelim_action_chunking} as self-supervision for online policy adaptation. Figure~\ref{fig:method_overview} illustrates the overall adaptation process. Across the shift boundary, \textbf{Transition Anchoring} provides a temporally aligned nominal-side action reference for correcting the policy under the shifted observation. As execution proceeds, the adapted policy is carried forward and the leftover supervision is repeatedly refreshed, yielding \textbf{Sequential Correction Propagation} along the execution trajectory. In deployment, a lightweight adaptation gate determines when these online updates begin.

\subsection{Leftover Supervision}
\label{sec:leftover_supervision}

At replan $r_t$ with $t \geq 1$, we denote the leftover trajectory from the preceding action chunk by
\begin{equation}
    \mathbf{Y}_t
    :=
    \mathbf{A}_{t-1}[E{:}H]
    =
    [\mathbf{a}_{t-1,E},\ldots,\mathbf{a}_{t-1,H-1}]
    \in \mathbb{R}^{L\times d}.
    \label{eq:leftover_target}
\end{equation}
By construction, $\mathbf{Y}_t$ and the prefix of the current action chunk, $\mathbf{A}_t[0{:}L]$, correspond to the same future control interval. More specifically, for each $j=0,\ldots,L-1$, the target action $\mathbf{Y}_{t,j}=\mathbf{a}_{t-1,E+j}$ and the current prediction $\mathbf{a}_{t,j}$ refer to the same future control step. We call this one-to-one correspondence \emph{temporal alignment}.

When adaptation is active, \systems{} treats $\mathbf{Y}_t$ as a detached action-space target for updating the policy conditioned on the current input $(o_t,s_t,g)$. Importantly, $\mathbf{Y}_t$ contains planned actions for the shared future control interval rather than actions that have already been executed. Thus, SALT obtains its supervision directly from predictions produced during closed-loop execution.

\subsection{Temporal Roles of Leftover Supervision}
\label{sec:temporal_roles}

\paragraph{Transition Anchoring.}
At the shift boundary, the leftover target comes from a chunk generated under the nominal visual condition:
\begin{equation}
    \mathbf{A}_{t^*-1}
    \sim
    \pi_{t^*-1}
    \bigl(
        \cdot
        \mid
        o_{t^*-1}^{\mathrm{nom}},
        s_{t^*-1},
        g
    \bigr),
    \qquad
    \mathbf{Y}_{t^*}
    :=
    \mathbf{A}_{t^*-1}[E{:}H].
    \label{eq:transition_anchor}
\end{equation}

If no adaptation has occurred before the shift boundary, then $\pi_{t^*-1}=\pi_{\mathrm{base}}$. In either case, $\mathbf{Y}_{t^*}$ is generated from a nominal visual observation, whereas the current policy input contains the shifted observation $o_{t^*}^{\mathrm{shift}}$.
Thus, at $t^*$, the leftover and the current policy prediction differ in their visual conditioning.

Crucially, this change in visual mismatch does not introduce a temporal mismatch. Although $\mathbf{Y}_{t^*}$ was predicted from the preceding observation, $\mathbf{Y}_{t^*}$ and $\mathbf{A}_{t^*}[0{:}L]$ correspond to the same future control interval starting at $t^*$.

\begin{wrapfigure}[16]{r}{0.30\linewidth}
    \vspace{-1.5em}
    \centering
    \includegraphics[width=\linewidth]{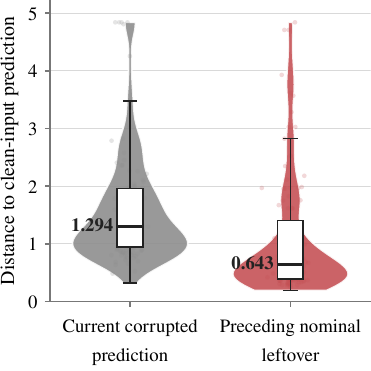}
    \captionsetup{font=small}
    \caption{\textbf{Leftover quality at transition.}
    Median 42-step $\ell_2$ distance to the clean-input prediction across
    1,000 pairs: 0.643 for the leftover and 1.294 for the corrupted-input
    prediction.}
    \label{fig:leftover_teacher_onset}
\end{wrapfigure}

Moreover, the leftover better preserves the nominal-side behavior of the
frozen policy.
Relative to the prediction that $\pi_{\mathrm{base}}$ produces from the clean
observation at the current state, the leftover is substantially closer than the
prediction produced from the current shifted observation
(median 42-step $\ell_2$ distance: 0.643 vs.\ 1.294 across 1,000 shift-boundary states;
Figure~\ref{fig:leftover_teacher_onset}).
Thus, despite having been generated one action chunk earlier, the leftover
retains a useful action-space reference for the behavior intended before the
visual shift.

We refer to this cross-boundary use of the leftover trajectory as
\emph{Transition Anchoring}.
When adaptation is active at the shift boundary, the update of
Equation~\ref{eq:adaptation_update} (Section~\ref{sec:online_pipeline}) takes
the shifted observation $o_{t^*}^{\mathrm{shift}}$ as the policy input and the
nominal-side leftover $\mathbf{Y}_{t^*}$ as the target. This update pulls the policy
under the shifted observation toward the plan formed before the shift.

\paragraph{Sequential Correction Propagation.}
After a cross-boundary correction is established, subsequent active replans $r_t$ with $t>t^*$ use refreshed leftover targets generated by the adapted policy under shifted observations:
\begin{equation}
    \mathbf{A}_{t-1}
    \sim
    \pi_{t-1}
    \bigl(
        \cdot
        \mid
        o_{t-1}^{\mathrm{shift}},
        s_{t-1},
        g
    \bigr),
    \qquad
    \mathbf{Y}_{t}
    =
    \mathbf{A}_{t-1}[E{:}H],
    \qquad
    t>t^*,
    \label{eq:sequential_correction}
\end{equation}
where $\pi_{t-1}$ already carries the updates made since activation. Each
such replan applies Equation~\ref{eq:adaptation_update} with this target,
generates $\mathbf{A}_t$ from the updated policy, and keeps its tail as
$\mathbf{Y}_{t+1}$. Thus, the transition target in Equation~\ref{eq:transition_anchor} originates from a policy prediction under a nominal observation, whereas subsequent targets are refreshed from the adapted policy operating under shifted observations. The adapted parameters are carried forward across these updates rather than reset to their pretrained initialization.

We refer to this cumulative process as \emph{Sequential Correction Propagation}. Propagation here does not imply that a single transition-time update globally corrects the policy on distant future observations. Instead, Transition Anchoring seeds a cross-boundary correction, while subsequent updates locally refine this correction as execution reaches new trajectory regions. By carrying the adapted policy forward and repeatedly refreshing the leftover target, \systems{} progressively carries the anchor-seeded correction forward along the execution trajectory. Section~\ref{sec:propagation} probes this carryover at a fixed target observation (Figure~\ref{fig:local_propagation_recovery}(b)) and evaluates its closed-loop benefit.

\subsection{Online Adaptation Pipeline}
\label{sec:online_pipeline}

Sections~\ref{sec:leftover_supervision} and~\ref{sec:temporal_roles} define the leftover supervision and temporal roles that constitute the \systems{} adaptation mechanism. We now describe how this mechanism is instantiated as an online deployment pipeline.

\paragraph{Adaptation gate.}
The shift boundary $t^*$ is unknown at deployment, so the pipeline uses a lightweight adaptation gate to determine when \systems{} begins updating.
At each replan $r_t$, we compare visual tokens $F_t$ with those from the preceding replan, $F_{t-1}$. Adaptation is activated when the resulting temporal feature change exceeds a threshold calibrated on a disjoint set of nominal trajectories. Once activated, adaptation remains active for the rest of the episode. 
The gate determines when the policy may change, helping to reduce average computation (Appendix~\ref{app:runtime}).
The feature-change score, calibration procedure, and activation statistics are provided in Appendix~\ref{app:implementation} and Appendix~\ref{app:detector_results}.



\paragraph{Leftover-supervised policy update.}
For the flow-based VLA backbones considered in this work, we instantiate the leftover-supervised update using each backbone's native flow-matching objective.

We write the policy at replan $r_t$ as $\pi_t \equiv \pi_{\theta,\phi_t}$, where $\theta$ denotes the frozen pretrained parameters and $\phi_t$ denotes the adaptive parameters. \systems{} optimizes only a lightweight set of adaptive parameters associated with selected action-expert projections. 

To supervise the $L$ aligned actions in $\mathbf{Y}_t$, we pad the target to the full chunk shape as $\widetilde{\mathbf{Y}}_t$ and define a binary mask $M$ over the aligned positions and valid action dimensions. In a standard flow-matching formulation, for $\tau\sim\mathcal{U}[0,1]$ and $\epsilon\sim\mathcal{N}(0,I)$, we construct
\[
    X_\tau
    =
    (1-\tau)\widetilde{\mathbf{Y}}_t
    +
    \tau\epsilon
\]
and write the leftover-supervised objective for input $(o_t,s_t,g)$ and target $\mathbf{Y}_t$ as
\begin{equation}
    \mathcal{L}(\phi;\,o_t,s_t,g,\mathbf{Y}_t)
    =
    \mathbb{E}_{\tau,\epsilon}
    \left[
    \left\|
    M \odot
    \left(
        v_{\theta,\phi}
        (X_\tau,\tau,o_t,s_t,g)
        -
        (\epsilon-\widetilde{\mathbf{Y}}_t)
    \right)
    \right\|_F^2
    \right].
    \label{eq:adaptation_loss}
\end{equation}


At each active replan, the adaptive parameters are updated by gradient descent on Equation~\ref{eq:adaptation_loss} starting from $\phi_{t-1}$, while $\mathbf{Y}_t$ remains detached:
\begin{equation}
    \phi_t
    =
    \phi_{t-1}
    -
    \alpha\,\nabla_{\phi}\mathcal{L}(\phi_{t-1};\,o_t,s_t,g,\mathbf{Y}_t),
    \label{eq:adaptation_update}
\end{equation}
where $\alpha$ is the learning rate. Exact parameterization, backbone-specific flow conventions, and optimization hyperparameters are provided in Appendix~\ref{app:implementation}.

\paragraph{Execution.}
At an inactive replan, the predicted chunk is executed as is. At an active replan, \systems{} updates the adaptive parameters and regenerates the action chunk from the same input before execution. The robot then executes $\mathbf{A}_t[0{:}E]$, and the regenerated chunk's tail becomes $\mathbf{Y}_{t+1}$.



\section{Experiments}
\label{sec:experiments}

\begin{table*}[!t]
\centering
\caption{
\textbf{Success rates (\%) under five persistent visual corruptions on LIBERO-10}. Avg. averages the five corrupted conditions. Clean is reported separately. Bold marks the best corrupted-condition result within each backbone.
}
\label{tab:main}
\label{tab:transfer}
\small
\renewcommand{\arraystretch}{1.15}
\setlength{\tabcolsep}{4pt}
\begin{tabularx}{\textwidth}{@{}lXrrrrrrr@{}}
\toprule
Backbone & Method & Clean & Motion & Gaussian & Zoom & Glass & Fog & Avg. \\
\midrule
\multirow{5}{*}{SmolVLA} & Frozen & 71.5 & 63.0 & 25.0 & 29.0 & 56.5 & 46.0 & 43.9 \\
 & ActMAD~\citep{mirza2023actmad} & 74.5 & 60.0 & 29.0 & 31.0 & 54.0 & 48.5 & 44.5 \\
 & BID~\citep{liu2025bidirectional} & 69.5 & 63.0 & 27.0 & 32.5 & 57.0 & 50.5 & 46.0 \\
 & Self-training & 71.0 & 61.0 & 23.5 & 26.5 & 55.5 & 48.0 & 42.9 \\
 & \cellcolor{ourscolor}\textbf{\systems{}} & \cellcolor{ourscolor}70.5 & \cellcolor{ourscolor}\textbf{68.5} & \cellcolor{ourscolor}\textbf{43.0} & \cellcolor{ourscolor}\textbf{42.0} & \cellcolor{ourscolor}\textbf{58.5} & \cellcolor{ourscolor}\textbf{54.0} & \cellcolor{ourscolor}\textbf{53.2} \\
\midrule
\multirow{5}{*}{GR00T N1.7} & Frozen & 81.5 & 76.5 & 46.0 & 53.0 & 40.0 & 78.0 & 58.7 \\
 & ActMAD~\citep{mirza2023actmad} & 82.0 & 79.0 & 49.5 & 55.0 & 41.5 & 77.0 & 60.4 \\
 & BID~\citep{liu2025bidirectional} & 82.0 & \textbf{79.5} & 53.0 & 60.0 & 44.0 & 79.0 & 63.1 \\
 & Self-training & 82.0 & 77.5 & 41.5 & 52.0 & 32.5 & \textbf{80.5} & 56.8 \\
 & \cellcolor{ourscolor}\textbf{\systems{}} & \cellcolor{ourscolor}82.0 & \cellcolor{ourscolor}78.5 & \cellcolor{ourscolor}\textbf{55.5} & \cellcolor{ourscolor}\textbf{63.5} & \cellcolor{ourscolor}\textbf{54.0} & \cellcolor{ourscolor}78.5 & \cellcolor{ourscolor}\textbf{66.0} \\
\bottomrule
\end{tabularx}
\end{table*}

\subsection{Experimental Setup}
\label{sec:experimental_setup}

We evaluate two VLA backbones, SmolVLA~\citep{shukor2025smolvla} and GR00T
N1.7~\citep{bjorck2025groot}, in simulation. The real-world evaluation uses
SmolVLA fine-tuned on the tabletop tasks and deployed on an SO-101 robot.
We compare \systems{} with the
frozen policy, ActMAD~\citep{mirza2023actmad}, BID~\citep{liu2025bidirectional},
and Self-training. ActMAD is a non-entropy-based test-time adaptation method
that matches activation statistics; BID uses leftover actions for guided
inference-time chunk selection. Self-training replaces the leftover target
with the detached current prediction while matching \systems{}'s gate and
update procedure. In both simulation and real-world trials, disruptions arise
during execution without providing their type or timing to the policy, and
all methods are evaluated under the same conditions within each setting.
Implementation details for \systems{} and the baselines are given in
Appendices~\ref{app:implementation} and~\ref{app:baselines}.

\subsection{Simulation-Based Evaluation}
\label{sec:simulation_setup}
\label{sec:main_results}
\label{sec:transfer}

\paragraph{Benchmark and visual disruptions.}
We evaluate all ten LIBERO-10 tasks~\citep{liu2023libero}, using 20 episodes
per task and condition. We apply five LIBERO-Plus sensor-noise
transformations: motion blur, Gaussian blur, zoom
blur, glass blur, and fog ~\citep{fei2026liberoplus}. Each is applied at severity 5 to both the
agent-view and wrist-camera images. Figure~\ref{fig:compare_noise} shows
examples of the five corruptions across severity levels. Each disruption
begins at a random point between 30\% and 70\% of a clean reference
rollout, rounded to a replan, and persists until the episode ends. Robot
state and language instruction remain unchanged. Methods with the same backbone share
initial states and shift boundaries. We report task success, with clean
performance shown separately and the five corrupted conditions weighted
equally in Avg. Appendix~\ref{app:episodeconstruction} gives the full
episode-construction protocol.

\paragraph{Main simulation results.}
Table~\ref{tab:main} shows that \systems{} increases the
corrupted-condition average from 43.9\% to 53.2\% on SmolVLA and from
58.7\% to 66.0\% on GR00T N1.7 relative to the frozen policies. It also
exceeds the strongest baseline average by 7.2 and 2.9 percentage points,
respectively. Relative to Frozen, success improves in all five corruption
categories on both backbones, while clean success remains within one
percentage point. The gains are largest where the frozen policies are
particularly vulnerable: Gaussian and zoom blur for SmolVLA, and
Gaussian, zoom, and glass blur for GR00T N1.7. Additional LIBERO suites and corruption severities are
reported in Appendix~\ref{app:additional_results}, and further deployment
details in Appendix~\ref{app:runtime}.

\FloatBarrier

\subsection{Real-World Robot Evaluation}
\label{sec:realworld}

\begin{figure*}[!t]
\centering
\includegraphics[width=\textwidth]{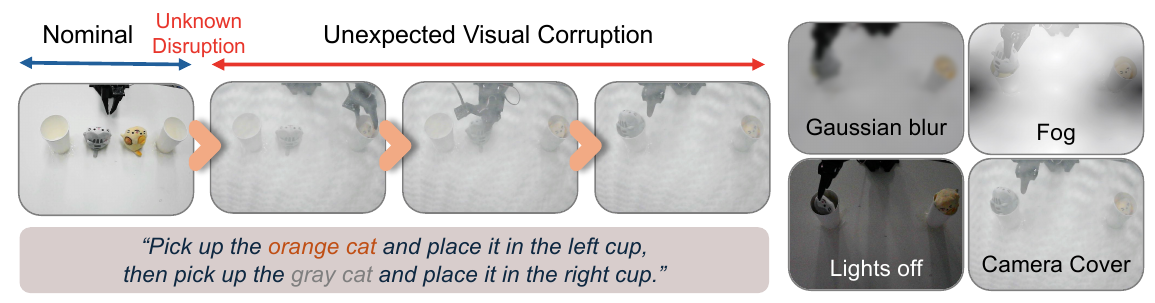}
\caption{\textbf{Representative real-world visual disruptions.}
Nominal and disrupted observations are shown for Two Cats in Cups.
Gaussian blur and fog are applied digitally to the camera stream; lights off
and camera cover are physical disturbances introduced during execution.}
\label{fig:visual_disruption}
\end{figure*}

\paragraph{Tasks and disruption conditions.}
We evaluate three multi-stage tabletop tasks: Two Cats in Cups (place the
orange and gray cats in the left and right cups), Drawer (place a block in a
drawer and close it), and Stack (stack the middle block on the right block,
then place the left block in a cup).
During execution, we introduce visual disruptions digitally by applying
Gaussian blur or fog to the camera streams, or physically by turning off the
room lights or covering the overhead camera with a mesh.
Figure~\ref{fig:visual_disruption} shows representative observations. Each
disruption starts at a fixed replan about 30\% into a successful nominal trial: replan
10 for Two Cats in Cups and Stack, and replan 6 for the shorter Drawer task.
Each task-condition pair uses 20 trials per method. Since the tasks are long-horizon with multiple sub-goals, we report a task-progress score in $[0,1]$ that gives partial credit for completed sub-goals, following common practice for distinguishing partial completion in long-horizon manipulation \citep{Zhang_2025_ICCV, snyder2026beyond}. Scoring and robot-protocol details are in Appendix~\ref{app:realrobot}.

\begin{figure*}[!t]
\centering
\includegraphics[width=\textwidth]{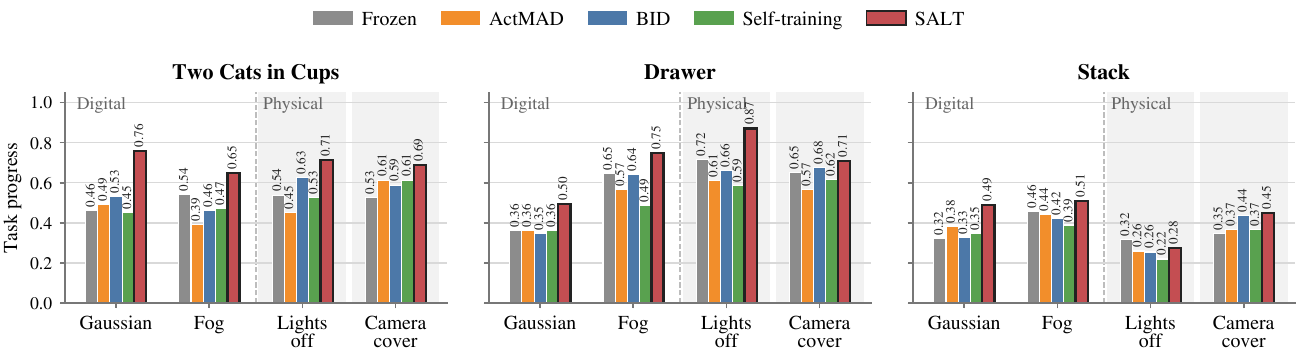}
\caption{\textbf{Real-world task-progress score under digital and
physical disruptions.} Each bar averages 20 trials per method. A score of 1.0 means the task was completed.}
\label{fig:realworld}
\end{figure*}

\paragraph{Main real-world results.}
Figure~\ref{fig:realworld} reports task progress under digital and
physical disruptions. Across the three tasks and four conditions,
\systems{} achieves the highest score in every cell except Stack under
lights off. In that condition no method preserves meaningful progress. Averaged over the three tasks, \systems{} raises task progress over the frozen policy from 0.46 to 0.61 under the digital disruptions and from 0.51 to 0.62 under the physical ones.
The second-best method varies with the task and condition, whereas
\systems{} ranks consistently at the top. Lights off and camera cover
change the physical scene and camera input rather than applying an image
operator, so the gains under these conditions show that the benefit
extends beyond simulated corruptions.

\begin{figure*}[t]
    \centering
    \begin{subfigure}[t]{0.32\textwidth}
        \centering
        \includegraphics[width=\linewidth]{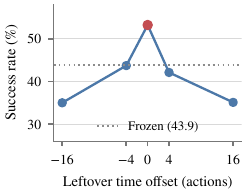}
        \caption{Temporal alignment.}
        \label{fig:analysis_temporal_alignment}
    \end{subfigure}\hfill
    \begin{subfigure}[t]{0.32\textwidth}
        \centering
        \includegraphics[width=\linewidth]{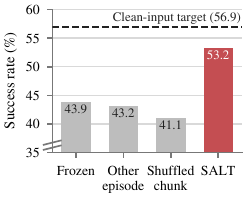}
        \caption{Target identity.}
        \label{fig:analysis_target_comparison}
    \end{subfigure}\hfill
    \begin{subfigure}[t]{0.34\textwidth}
        \centering
        \includegraphics[width=\linewidth]{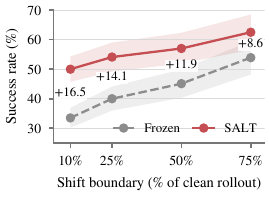}
        \caption{Shift timing.}
        \label{fig:analysis_onset_sensitivity}
    \end{subfigure}
    \caption{\textbf{Leftover target properties and shift timing on SmolVLA.}
    \textbf{(a) Temporal alignment.} Shifting the leftover away from its
    aligned position at offset 0 reduces success to or below Frozen.
    \textbf{(b) Target identity.} Another episode's plan and a time-shuffled
    previous chunk remain at or below Frozen, while the aligned leftover
    approaches the clean-input prediction target.
    \textbf{(c) Shift timing.} Success when the shift boundary is fixed at
    four points of the clean rollout. Shaded bands show 95\% confidence
    intervals. Exact values for (c) are reported in Table~\ref{tab:shift_timing}.}
    \label{fig:leftover_controls}
\end{figure*}

\FloatBarrier

\section{Analysis}
\label{sec:analysis}

\subsection{Transition Anchoring: Why Is the Leftover a Good Target?}
\label{sec:anchoring}
\label{sec:ablation_study}

\begin{wrapfigure}{r}{0.33\linewidth}
    \vspace{-1.0em}
    \centering
    \includegraphics[width=\linewidth]{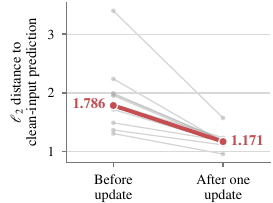}
    \captionsetup{font=small}
    \caption{\textbf{One update narrows the clean-plan gap} (SmolVLA,
    Gaussian blur; thin: tasks, thick: pooled).}
    \label{fig:leftover_gap_closure}
    \vspace{-1.0em}
\end{wrapfigure}

At the shift boundary, the leftover was planned from a nominal observation
yet refers to the same future interval as the current prediction
(Section~\ref{sec:temporal_roles}), and learning from it moves the corrupted
prediction toward the clean plan: under Gaussian blur, one update lowers the
median $\ell_2$ distance to the clean-input prediction at the same state from
1.786 to 1.171 on SmolVLA ($n=200$; Figure~\ref{fig:leftover_gap_closure}) and from 0.528 to 0.335 on
GR00T N1.7 (Appendix~\ref{app:mechanism_diagnostics}).

As shown in Figure~\ref{fig:leftover_controls}(a), shifting the leftover by
as few as four actions reduces success from 53.2\% to
the level of the frozen policy. This shows that the leftover is useful when
its actions remain temporally aligned with the current future control
interval. Figure~\ref{fig:leftover_controls}(b) asks whether another nominal
action target can provide the same benefit. Replacing the leftover with
another episode's plan for the same interval or a time-shuffled previous
chunk leaves performance at or below the frozen level, whereas the aligned
leftover approaches the performance of the clean-input prediction target.
This shows that the target must also remain tied to the current trajectory
rather than simply being a nominal action sequence. When the corruption is
present from the first replan, no leftover has been generated from a nominal
observation and adaptation leaves success nearly unchanged. We report this
setting in Appendix~\ref{app:scope_details}. Together, these results support
the leftover as a temporally aligned nominal-side action reference at the
visual transition.

\subsection{Sequential Correction Propagation: Benefits of Continued Updates}
\label{sec:propagation}
\label{sec:local_propagation}

\begin{figure}[t]
\centering
\begin{subfigure}[b]{0.59\linewidth}
    \centering
    \small
\setlength{\tabcolsep}{4pt}
\begin{tabular}{@{}lcccr@{}}
\toprule
& Anchoring & \multicolumn{2}{c}{Propagation} & \\
\cmidrule(lr){2-2}\cmidrule(lr){3-4}
Variant & \shortstack{Update at\\activation} & \shortstack{Later\\updates} & \shortstack{Adapted\\state kept} & Success \\
\midrule
Frozen & \no & \no & \no & 43.9 \\
One-time adaptation & \yes & \no & \no & 45.5 \\
Without anchoring & \no & \yes & \yes & 48.9 \\
Per-replan reset & \yes & \yes & \no & 45.0 \\
\rowcolor{ourscolor}
\textbf{\systems{}} & \yes & \yes & \yes & \textbf{53.2} \\
\bottomrule
\end{tabular}
    \vspace{0.4em}
    \caption{Ablations.}
\end{subfigure}
\hfill
\begin{subfigure}[b]{0.38\linewidth}
    \centering
    \includegraphics[width=\linewidth]{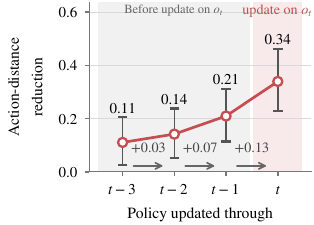}
    \caption{Sequential correction.}
\end{subfigure}
\caption{\textbf{Roles and propagation of leftover-supervised updates on
SmolVLA.} \textbf{(a)} Ablations of the two roles: success over
the 1,000 episodes of Table~\ref{tab:main}; without anchoring skips only the
update at the shift boundary. \textbf{(b)} Sequential correction: normalized
$\ell_2$-distance reduction for policies updated through replan $t-3,\dots,t$
(40 replayed clean trajectories, five corruptions; 95\% bootstrap intervals).
All policies are evaluated on the same $o_t$.}
\label{fig:local_propagation_recovery}
\end{figure}

Figure~\ref{fig:local_propagation_recovery}(a) examines how Sequential
Correction Propagation complements the transition-time correction established
by Transition Anchoring. Full \systems{} reaches 53.2\% success. In contrast,
a single update at activation reaches 45.5\%, and continued updates that
restart from the pretrained weights at each replan reach 45.0\%. Removing
the transition-time update also lowers performance. These ablations show
that both Transition Anchoring and Sequential Correction Propagation are
needed to achieve the full improvement of \systems{}.

The action-space analysis in Figure~\ref{fig:local_propagation_recovery}(b)
shows how Sequential Correction Propagation carries the correction forward
to observations that have not yet been seen. On the same later observation
$o_t$, policies updated through $t-3$, $t-2$, and $t-1$ already reduce
the distance to the clean-input prediction, even though none has been updated
on $o_t$. The reduction becomes larger as the adapted state is carried closer
to $t$, and updating on $o_t$ provides an additional local correction. This
shows that corrections from earlier replans can transfer to future
observations, while repeated updates continue to refresh and strengthen that
effect along the execution trajectory.

By refreshing the leftover target and carrying the adapted state forward at
each replan, \systems{} continues correcting its predictions as the robot
reaches new observations. \systems{} also shows larger improvements over Frozen when disruptions
begin earlier in the rollout, despite the corruption affecting a larger
fraction of the rollout, as shown in Figure~\ref{fig:leftover_controls}(c). Leftover supervision
also helps when the corruption increases gradually or clears after several
replans, with additional results reported in Appendix~\ref{app:scope_details}.

\section{Conclusion}
\label{sec:conclusion}

In this work, we study online adaptation of VLAs to visual disruptions whose
type and timing are unknown to the deployed policy. We propose \systems{}, which
uses the unexecuted leftover trajectory of the previous action chunk as
self-supervision. Because the leftover and the next chunk's prefix cover the
same future control interval, their temporal alignment provides action-space
supervision within the ongoing rollout. 
Unlike prior work that reuses overlapping action chunks to improve execution with the policy parameters fixed, \systems{} uses this overlap to update the policy itself. 
The leftover supervision plays two complementary roles, Transition Anchoring across the shift
boundary and Sequential Correction Propagation as execution continues. 
Across simulation and a real robot, \systems{} improves performance under visual disruptions.
More broadly, our results suggest that the temporal overlap in action-chunked execution is an underused resource for test-time adaptation of robot policies.

\bibliography{iclr2027_conference}
\bibliographystyle{iclr2027_conference}

\clearpage
\appendix

\section{Evaluation Details}
\label{app:episodeconstruction}
\label{app:protocol}

\paragraph{Episodes.}
Each LIBERO-10 task is evaluated from 20 initial states per condition. The
corruption begins at a random point between 30\% and 70\% of a clean rollout
of the frozen policy from the same initial state, rounded to a replan, and
persists until the episode ends. All methods of a backbone share the initial
states and shift boundaries. The clean rollouts only schedule the shift
boundaries; they are never
given to \systems{}.

\begin{figure}[t]
\centering
\includegraphics[width=0.95\linewidth]{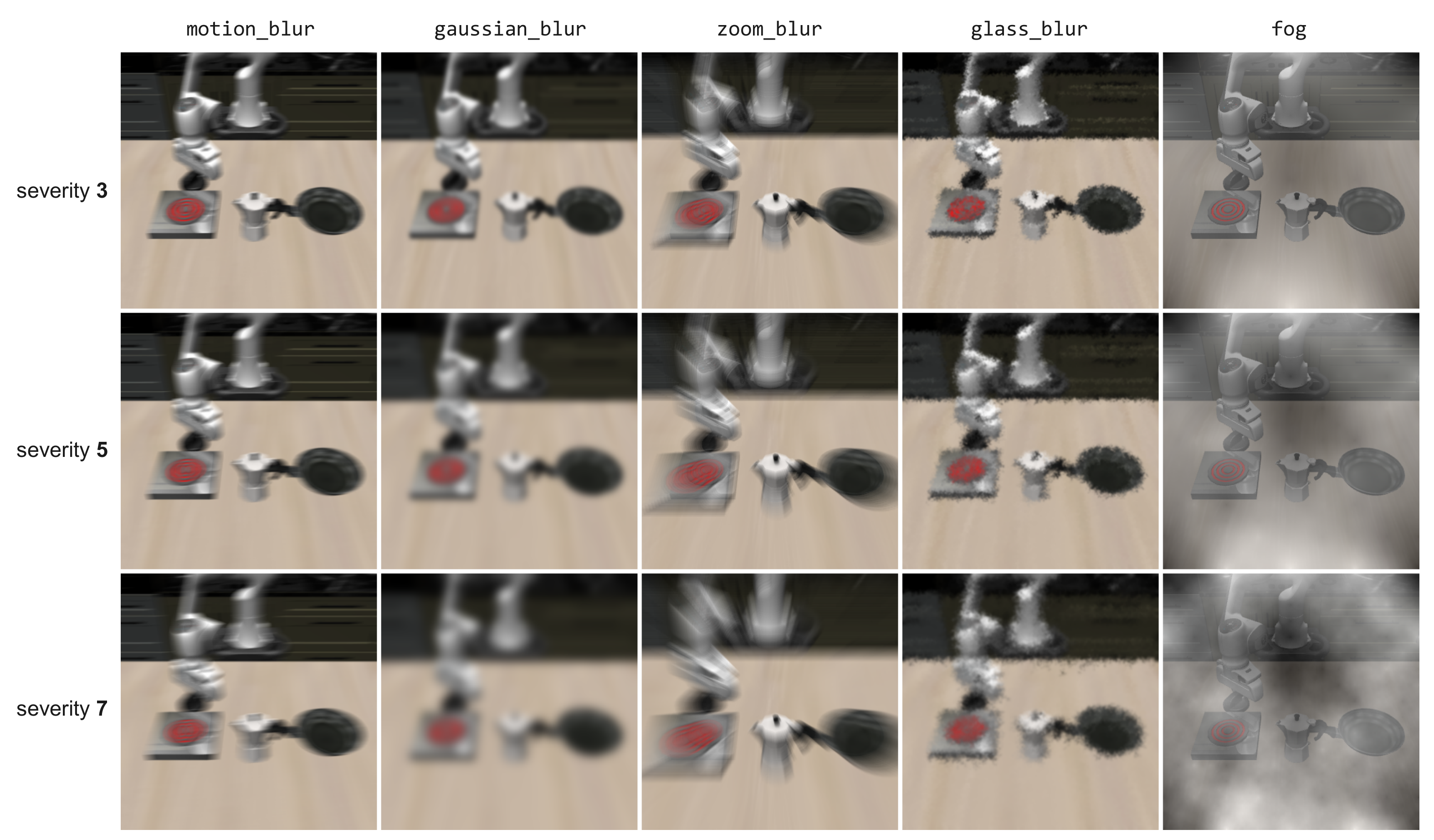}
\caption{\textbf{Visual corruption examples by severity.} The five
LIBERO-Plus corruption families at severity levels 3, 5, and 7.}
\label{fig:compare_noise}
\end{figure}

\paragraph{Corruptions.}
The five corruptions and their severity levels follow the sensor-noise
perturbations of LIBERO-Plus~\citep{fei2026liberoplus}
(Table~\ref{tab:corruption_params}); level 5 is used unless stated otherwise.
Unlike LIBERO-Plus, which corrupts only the agent-view image, we corrupt both
the agent-view and wrist images, since disruptions of the scene, such as a
change in lighting, reach every camera. Results under the default LIBERO-Plus
setting, which corrupts only the agent view, are in
Table~\ref{tab:agentview_only}. The real-robot camera-cover condition
covers the case in which only one camera is disrupted
(Appendix~\ref{app:realrobot}). Figure~\ref{fig:compare_noise} visualizes the
five corruption families across the evaluated severity levels.

\begin{table}[ht]
\centering
\caption{\textbf{Corruption parameters by severity level}, taken from the
LIBERO-Plus tables. Zoom blur averages the image over zoom factors from 1.0
to the listed maximum in the listed step.}
\label{tab:corruption_params}
\small
\setlength{\tabcolsep}{4pt}
\begin{tabular}{@{}llccc@{}}
\toprule
Corruption & Parameters & Level 3 & Level 5 & Level 7 \\
\midrule
Motion blur & radius, $\sigma$ (px) & 10, 4 & 15, 6 & 20, 10 \\
Gaussian blur & $\sigma$ (px) & 3 & 5 & 7 \\
Zoom blur & max.\ zoom (step) & 1.20 (0.02) & 1.30 (0.03) & 1.40 (0.01) \\
Glass blur & $\sigma$, displacement, iterations & 0.9, 2, 3 & 1.1, 3, 2 & 1.5, 4, 2 \\
Fog & strength, decay & 1.5, 2.5 & 2.5, 2.0 & 3.5, 1.6 \\
\bottomrule
\end{tabular}
\end{table}

\paragraph{Calibration.}
For each backbone, the adaptation gate is calibrated on 20 nominal
trajectories from five tasks, which are disjoint from the evaluation episodes
and contain no corrupted frames. The update hyperparameters in
Table~\ref{tab:implementation} are shared by both backbones, all corruptions,
the other LIBERO suites, and the real robot.

\paragraph{Statistics.}
Each corruption has 200 episodes, and the corruption average weights the five
families equally. Because episodes are paired across methods, $p$-values are
two-sided exact binomial tests on the episodes that only one of two methods
solves. Intervals for action-space diagnostics resample task--episode
clusters.

\section{Method Details}
\label{app:implementation}

\subsection{Online Algorithm}
\label{app:algorithm}

At the start of every episode, the adaptive parameters are set to zero, the
optimizer state is cleared, and the gate is inactive. At each replan:
\begin{enumerate}
 \item Obtain the current observation and a provisional action chunk together
 with the interface features.
 \item If the gate is inactive, compute the feature-change score
 (Equation~\ref{eq:method_detection_score}) and latch the gate when it exceeds
 the calibrated threshold. Once latched, the gate stays active for the rest of
 the episode.
 \item If the gate is active and a previous chunk exists, form the detached
 leftover target from its unexecuted tail, take five Adam steps on
 Equation~\ref{eq:adaptation_loss}, and regenerate the chunk with the updated policy.
 \item Otherwise, use the provisional chunk.
 \item Execute the first $E$ actions and keep the chunk for the next target.
\end{enumerate}
The update is synchronous: execution waits for the regenerated chunk.

\subsection{Updated Parameters}
In Equation~\ref{eq:adaptation_loss}, $\phi$ consists of log-scale vectors
$\boldsymbol\beta$ on selected attention key projections. This
parameterization builds on singular-value fine-tuning (SVF), which updates
singular values while fixing the singular vectors \citep{sun2022singular}. It is
also a diagonal case of SVFT, which more generally learns sparse coefficients
in the singular-vector basis \citep{lingam2024svft}. We use bounded
multiplicative scales:
\begin{equation}
\begin{aligned}
 W_0 &= U\operatorname{diag}(\boldsymbol\sigma)V^\top,\\
 W(\boldsymbol\beta)
 &= U\operatorname{diag}(\boldsymbol\sigma\odot
                  \exp(\boldsymbol\beta))V^\top,\\
 \boldsymbol\beta^{(0)}&=\mathbf 0,\qquad
 \|\boldsymbol\beta\|_\infty\leq\eta.
\end{aligned}
\label{eq:method_update_parameters}
\end{equation}

The elementwise exponential makes $\boldsymbol\beta=\mathbf 0$ recover
$W_0$. Clipping $\|\boldsymbol\beta\|_\infty\leq\eta$ bounds each singular
value within $[e^{-\eta}\sigma_i,e^\eta\sigma_i]$ while keeping the
singular directions fixed. The decomposition is computed once when the
gates are attached.
The learning rate is set to $2\eta/K$ for $K$ optimizer steps, so that the
bound is reachable within one update. The target is stored without gradients,
and the masked squared error is averaged over the supervised entries.
Table~\ref{tab:implementation} lists the model-specific settings.

\begin{table}[ht]
\centering
\caption{\textbf{Model-specific implementation.}}
\label{tab:implementation}
\small
\begin{tabularx}{\linewidth}{@{}Xcc@{}}
\toprule
Setting & SmolVLA & GR00T N1.7 \\
\midrule
Checkpoint & SmolVLA fine-tuned on LIBERO & GR00T N1.7 LIBERO-10 \\
Chunk $H$ / executed prefix $E$ / overlap $L$ & $50/8/42$ & $16/8/8$ \\
Adapted projections & \makecell{keys of the 8 cross-attention\\layers of the action expert} & \makecell{keys of the 16 cross-attention\\blocks of the action-head DiT} \\
Trainable scales & 2,560 & 24,576 \\
Bound $\eta$ & 0.15 & 0.15 \\
Optimizer & Adam, lr 0.06, 5 steps & Adam, lr 0.06, 5 steps \\
Noise--time draws per step & \makecell{8, flow times evenly\\spaced on $[0.001,1]$} & \makecell{8, native beta\\time sampling} \\
Supervised action dimensions & 7 & 7 \\
Reset & every episode & every episode \\
\bottomrule
\end{tabularx}
\end{table}

\subsection{GR00T Flow Convention}

The main-text flow equation is written in the SmolVLA convention, with the
action target at $\tau=0$, Gaussian noise at $\tau=1$, and target velocity
$\epsilon-\widetilde{Y}_t$. GR00T retains its native reverse-time convention.
For flow time $\tau'$ drawn from the pretrained action head's native time
distribution, it constructs
\begin{equation}
    X_{\tau'}=(1-\tau')\epsilon+\tau'\widetilde{Y}_t,
    \qquad
    U_{\tau'}=\widetilde{Y}_t-\epsilon.
    \label{eq:groot_flow_convention}
\end{equation}
Thus, under $\tau'=1-\tau$, the two conventions traverse the same linear
noise--action path with opposite velocity directions. Our GR00T
implementation invokes the native action-head objective, including its
beta-based time sampling and timestep discretization. We replicate the
target and mask across eight noise--time draws at each optimizer step.

\subsection{Adaptation Gate}

Let $F_r\in\mathbb{R}^{N\times D}$ denote the interface tokens from the
provisional pass at replan $r$. The temporal cosine change of token $i$ is
\begin{equation}
    d_{r,i}
    =
    1-
    \frac{\langle F_{r-1,i},F_{r,i}\rangle}
    {\|F_{r-1,i}\|_2\,\|F_{r,i}\|_2}.
    \label{eq:method_token_change}
\end{equation}
Calibration supplies token-wise robust centers $m_i$ and scales $s_i$, a
shared scale floor, and the threshold. \systems{} aggregates standardized
one-sided changes as
\begin{equation}
    S_r
    =
    \frac{1}{N}\sum_{i=1}^{N}
    \operatorname{clip}
    \left(
        \frac{d_{r,i}-m_i}{s_i},\,0,\,\kappa
    \right).
    \label{eq:method_detection_score}
\end{equation}
We take $m_i$ as the median and $s_i$ as $1.4826$ times the median absolute
deviation of the nominal changes at token $i$, and floor every $s_i$ at a
quarter of the median scale, so that a token which barely moves under nominal
conditions cannot turn a small change into a large score. The score cap is
$\kappa=10$. The threshold is cross-fitted by leaving out one calibration task
at a time: the statistics are refit on the remaining tasks, every held-out
nominal episode contributes its maximum score, and the threshold is $0.9$
times the largest of those maxima. The deployed $m_i,s_i$ are then refit on
all calibration episodes. The detector reads the agent-view stream for
backbones whose interface is one tensor per camera, and the leading vision
block of the sequence for backbones that emit vision and language fused;
token norms are floored before the cosine. The same rule, floor, and cap are
used for every model and corruption; only the threshold is recomputed per
model.

\section{Comparison Methods}
\label{app:baselines}

\paragraph{ActMAD.}
ActMAD~\citep{mirza2023actmad} aligns the means and variances of test-time
activations with statistics of nominal data. For SmolVLA, the statistics
cover the outputs of all 12 vision-encoder layers and are computed from 1,888
nominal frames of initial states 20--23 of all ten tasks. At every replan,
from the start of the episode and without a gate, the vision encoder takes
one SGD step (learning rate $2.5\times10^{-4}$, momentum 0.9, weight decay
$5\times10^{-4}$) on a batch of the eight most recent frames. For GR00T N1.7,
the statistics cover all 24 blocks of the vision transformer, computed from
1,593 nominal frames of the same initial states, and the update uses a
learning rate of $2.5\times10^{-6}$ with float32 master weights; the
SmolVLA rate made the GR00T policy fail on every clean episode.

\paragraph{BID.}
BID~\citep{liu2025bidirectional} samples $N=16$ candidate chunks at every
replan and selects one using backward coherence with the previous chunk
(decay $\rho=0.5$) and forward contrast against samples from a strong and a
weak policy (mode size $K=3$). The weak policy is an
early checkpoint of the same model: SmolVLA fine-tuned on LIBERO for 2,500
steps, and GR00T N1.7 fine-tuned for 1,000 of 10,000 steps. BID therefore
uses a second model and a larger sampling budget than the frozen policy.

\paragraph{Self-training.}
This control replaces the leftover target with the detached chunk that the
policy has just produced from the current observation, supervising all $H$
positions. The observation, updated parameters, optimizer, number of update
steps, and gate are matched to \systems{}.

\section{Additional Results}
\label{app:additional_results}
\label{app:full}
\label{app:robustness}

\paragraph{Other LIBERO suites.}
Table~\ref{tab:libero_suites} applies the same protocol to LIBERO-Spatial,
LIBERO-Object, and LIBERO-Goal, each with ten tasks and 20 episodes per task
and corruption. Each suite has its own gate threshold, calibrated with the
rule of Appendix~\ref{app:protocol} on nominal trajectories of that suite.
Averaged over the three suites, \systems{} raises SmolVLA from 74.3\% to
79.4\% and GR00T N1.7 from 87.6\% to 89.8\%. On SmolVLA, clean success is
unchanged on LIBERO-Object (93.5\%) and LIBERO-Goal (87.0\%) and moves from
80.5\% to 79.0\% on LIBERO-Spatial. The gains concentrate where the frozen
policy loses most; on GR00T LIBERO-Spatial, whose frozen success is already
within four points of clean, \systems{} is level on average and lowers
Gaussian-blur success.

\begin{figure}[t]
\centering
\includegraphics[width=\linewidth]{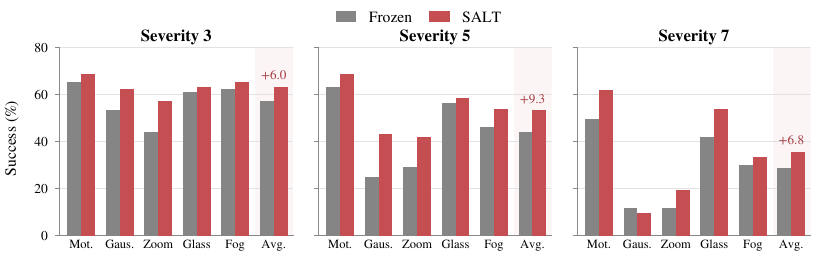}
\caption{\textbf{SmolVLA success rate (\%) on LIBERO-10 across
corruption severity.} Frozen and \systems{} are evaluated on 200 episodes
per corruption and severity. Mot.\ and Gaus.\ abbreviate motion and
Gaussian blur; Avg.\ excludes clean episodes. Numbers above Avg.\ are
\systems{} gains in percentage points.}
\label{fig:severity_sweep}
\end{figure}

\paragraph{Corruption severity.}
Figure~\ref{fig:severity_sweep} shows SmolVLA success on LIBERO-10 at
severity levels 3, 5, and 7 (Table~\ref{tab:corruption_params}). The
severity-5 values correspond to the main results in Table~\ref{tab:main}.

\begin{table}[ht]
\centering
\caption{\textbf{Agent-view-only corruption} (SmolVLA, LIBERO-10, success
\%). The wrist image stays clean. 200 episodes per cell.}
\label{tab:agentview_only}
\small
\setlength{\tabcolsep}{4pt}
\begin{tabular}{@{}lrrrrrr@{}}
\toprule
Method & Motion & Gaussian & Zoom & Glass & Fog & Avg. \\
\midrule
Frozen & 70.0 & 43.0 & 39.0 & 65.5 & 52.5 & 54.0 \\
\systems{} & 70.5 & 57.0 & 54.0 & 64.5 & 56.0 & \textbf{60.4} \\
\bottomrule
\end{tabular}
\end{table}

\begin{table*}[t]
\centering
\caption{\textbf{Results across LIBERO suites.} Success rate (\%) under
persistent severity-5 corruption, 1,000 episodes per suite and method.}
\label{tab:libero_suites}
\small
\setlength{\tabcolsep}{5pt}
\begin{tabular}{@{}lllrrrrrr@{}}
\toprule
Backbone & Suite & Method & Motion & Gaussian & Zoom & Glass & Fog & Avg. \\
\midrule
\multirow{8}{*}{SmolVLA}
 & \multirow{2}{*}{Spatial} & Frozen & 74.5 & 56.5 & 48.0 & 75.0 & \textbf{74.5} & 65.7 \\
 & & \systems{} & \textbf{75.5} & \textbf{70.5} & \textbf{68.0} & \textbf{75.5} & 72.0 & \textbf{72.3} \\
 \cmidrule(l){2-9}
 & \multirow{2}{*}{Object} & Frozen & \textbf{93.0} & 75.5 & 67.5 & \textbf{92.5} & \textbf{92.5} & 84.2 \\
 & & \systems{} & 91.5 & \textbf{88.0} & \textbf{86.5} & 91.0 & 89.5 & \textbf{89.3} \\
 \cmidrule(l){2-9}
 & \multirow{2}{*}{Goal} & Frozen & 80.5 & 67.5 & 57.5 & 79.5 & \textbf{80.5} & 73.1 \\
 & & \systems{} & \textbf{83.0} & 67.5 & \textbf{72.0} & \textbf{81.0} & 79.5 & \textbf{76.6} \\
 \cmidrule(l){2-9}
 & \multirow{2}{*}{Long (LIBERO-10)} & Frozen & 63.0 & 25.0 & 29.0 & 56.5 & 46.0 & 43.9 \\
 & & \systems{} & \textbf{68.5} & \textbf{43.0} & \textbf{42.0} & \textbf{58.5} & \textbf{54.0} & \textbf{53.2} \\
\midrule
\multirow{8}{*}{GR00T N1.7}
 & \multirow{2}{*}{Spatial} & Frozen & 91.0 & \textbf{87.0} & 86.0 & 88.5 & 90.5 & 88.6 \\
 & & \systems{} & \textbf{92.5} & 81.0 & \textbf{90.5} & \textbf{89.0} & \textbf{91.0} & \textbf{88.8} \\
 \cmidrule(l){2-9}
 & \multirow{2}{*}{Object} & Frozen & 94.5 & 88.0 & 92.0 & 92.0 & 94.5 & 92.2 \\
 & & \systems{} & \textbf{95.0} & \textbf{91.5} & \textbf{93.0} & \textbf{93.0} & 94.5 & \textbf{93.4} \\
 \cmidrule(l){2-9}
 & \multirow{2}{*}{Goal} & Frozen & 89.0 & 71.0 & 82.0 & 74.0 & 93.5 & 81.9 \\
 & & \systems{} & 89.0 & \textbf{81.5} & \textbf{86.5} & \textbf{85.0} & \textbf{94.0} & \textbf{87.2} \\
 \cmidrule(l){2-9}
 & \multirow{2}{*}{Long (LIBERO-10)} & Frozen & 76.5 & 46.0 & 53.0 & 40.0 & 78.0 & 58.7 \\
 & & \systems{} & \textbf{78.5} & \textbf{55.5} & \textbf{63.5} & \textbf{54.0} & \textbf{78.5} & \textbf{66.0} \\
\bottomrule
\end{tabular}
\end{table*}

\section{Real-World Protocol}
\label{app:realrobot}

\begin{figure*}[t]
    \centering
    \includegraphics[width=0.92\textwidth]{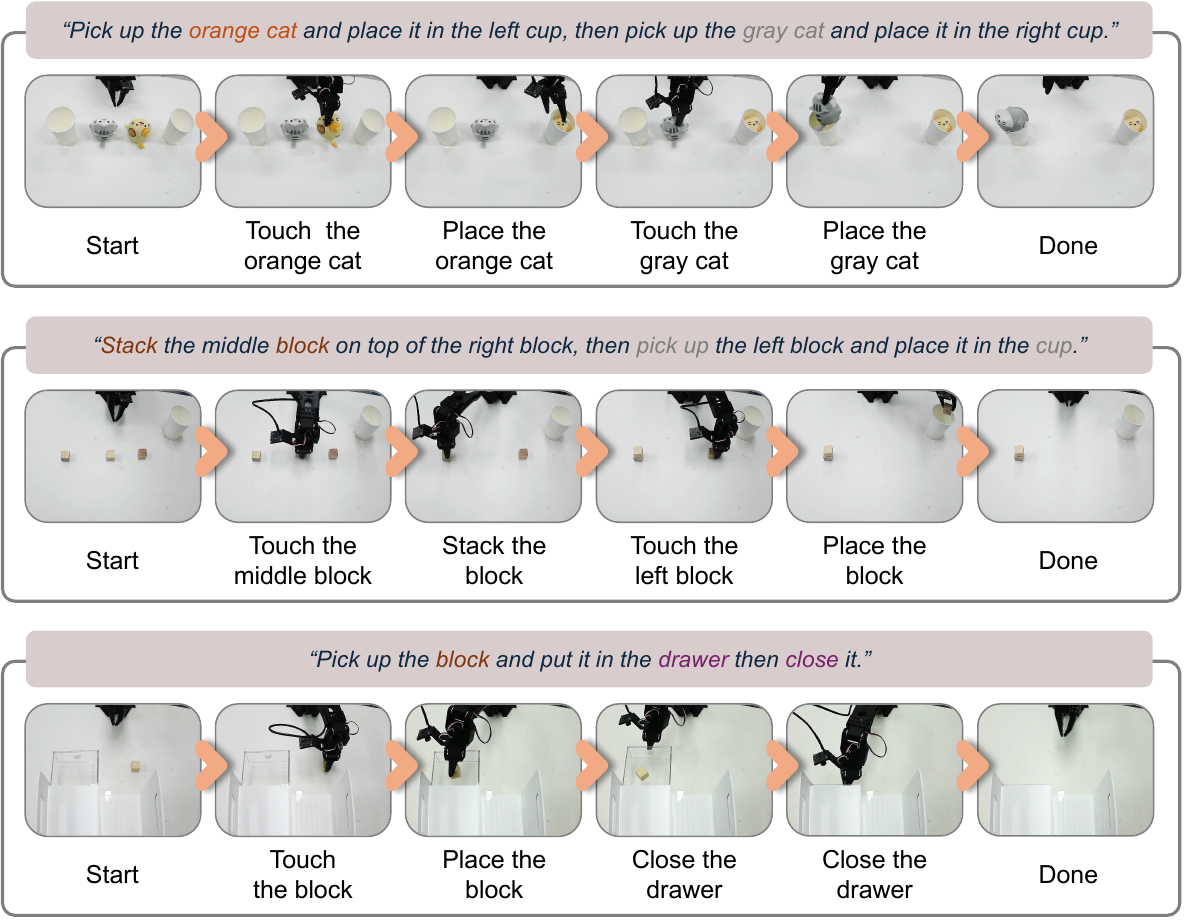}
    \caption{\textbf{Real-world tabletop tasks.} From top to bottom:
    \emph{Two Cats in Cups} places the orange and gray cats in the left and
    right cups, respectively; \emph{Stack} places the middle block on the
    right block, then the left block in a cup; and \emph{Drawer} places a
    block in the drawer and closes it. Each row shows successive stages
    from the starting scene to task completion.}
    \label{fig:realworld_tasks}
\end{figure*}

\paragraph{Tasks and policy.}
We evaluate three multi-stage tabletop tasks:
\begin{itemize}
    \item \textbf{Two Cats in Cups}: ``Pick up the orange cat and place it in
    the left cup, then pick up the gray cat and place it in the right cup.''
    \item \textbf{Drawer}: ``Pick up the block and put it in the drawer then
    close it.''
    \item \textbf{Stack}: ``Stack the middle block on top of the right block,
    then pick up the left block and place it in the cup.''
\end{itemize}
Figure~\ref{fig:realworld_tasks} shows the stages of these tasks.

The policy is SmolVLA fine-tuned on demonstrations of the tabletop tasks with
the vision encoder frozen. It reads an overhead and a wrist camera together
with the joint state, and the arm is controlled at 30~Hz. Each replan predicts
a 50-action chunk and executes its first 25 actions, so the leftover contains
25 actions. The gate is calibrated on nominal robot rollouts that are not
part of the evaluation, and the adaptation settings match simulation.

\paragraph{Disruptions.}
Each disruption begins at a fixed replan about 30\% into a successful nominal
trial (median length 35 replans on Two Cats in Cups, 34 on Stack, and 20 on
Drawer): replan 10 for Two Cats in Cups and Stack, and replan 6 for Drawer.
Gaussian blur and fog are applied at
severity 5 to both camera streams before inference, using the same
implementation as in simulation. For the physical disruptions, the operator
switches off the room lights or places a mesh in front of the overhead camera
lens; execution pauses for three seconds while the disturbance is applied.

\paragraph{Scoring.}
The task-progress score gives partial credit for completed sub-goals. For Two Cats in Cups,
touching the first cat scores 0.25, placing it in its cup 0.5, touching the
second cat 0.75, and placing both 1.0. For Drawer, touching the block,
placing it in the drawer, and closing the drawer contribute 0.3, 0.3, and
0.4. For Stack, touching the middle block, stacking it, touching the left
block, and placing it in the cup contribute 0.2, 0.3, 0.2, and 0.3. A trial
is a success when it reaches a score of 1.0.
Table~\ref{tab:realworld_success} reports binary success counts. The frozen
policy succeeds in 16 of 20 clean trials on Two Cats in Cups, 10 of 20 on
Drawer, and 4 of 20 on Stack. Binary success under disruption is rare for
every method on Stack, so the task-progress score in Figure~\ref{fig:realworld}
is the primary measure.

\begin{table*}[!htbp]
\centering
\caption{\textbf{Real-world success counts.} Successful trials out of 20 per
task, condition, and method.}
\label{tab:realworld_success}
\small
\renewcommand{\arraystretch}{1.08}
\setlength{\tabcolsep}{3pt}
\resizebox{\textwidth}{!}{%
\begin{tabular}{@{}l*{12}{c}@{}}
\toprule
& \multicolumn{4}{c}{Two Cats in Cups} & \multicolumn{4}{c}{Drawer}
& \multicolumn{4}{c}{Stack} \\
\cmidrule(lr){2-5}\cmidrule(lr){6-9}\cmidrule(l){10-13}
Method & Gaussian & Fog & Lights off & Camera cover
& Gaussian & Fog & Lights off & Camera cover
& Gaussian & Fog & Lights off & Camera cover \\
\midrule
Frozen & 2 & 0 & 1 & 1 & 0 & 3 & 9 & 5 & 0 & 0 & 0 & 0 \\
ActMAD & 3 & 0 & 0 & 3 & 0 & 4 & 8 & 4 & 0 & 1 & 0 & 0 \\
BID & 1 & 1 & 2 & 2 & 0 & 8 & 8 & \textbf{9} & 0 & 0 & 0 & 1 \\
Self-training & 1 & 0 & 1 & 2 & 0 & 3 & 7 & 4 & 0 & 0 & 0 & 0 \\
\rowcolor{ourscolor}
\textbf{\systems{}} & \textbf{7} & 0 & \textbf{3} & \textbf{5} & 0 & \textbf{9} & \textbf{14} & 8 & 0 & 0 & 0 & 1 \\
\bottomrule
\end{tabular}%
}
\end{table*}

\section{Role of the Adaptation Gate}
\label{app:gate_role}

The adaptation gate determines when \systems{} begins updating and is
separate from the leftover-supervision mechanism itself.
Table~\ref{tab:components_detail} isolates its role by changing only when the
same leftover-supervised update starts. On SmolVLA, adapting at every replan
without the gate reaches a comparable corrupted-condition average (54.5\%
versus 53.2\% with the gate) and clean success (69.5\% versus 70.5\%). The gate
mainly reduces computation. Because deployment does not announce whether or
when a disruption will arrive, \systems{} without the gate must adapt at
every replan. It updates the policy on about 98\% of replans under both
nominal and corrupted inputs, at $4.2\times$ the frozen policy's average
per-replan latency in either condition (Table~\ref{tab:latency_methods}).
With the gate, the detector adds 9~ms to a replan without an update, and only
13\% of nominal replans trigger an update, so the average per-replan latency
is $1.5\times$ the frozen policy's under nominal input and $2.9\times$ under
persistent corruption (measurement details in Appendix~\ref{app:runtime}).
Overall, the gate serves as a lightweight deployment mechanism that keeps
\systems{} near the frozen policy's cost whenever the input is nominal,
rather than as a precise disruption detector.

\begin{table*}[t]
\centering
\caption{\textbf{When \systems{} starts adapting} (SmolVLA, LIBERO-10, success \%).
Every row except Never runs the same leftover-supervised update; only its
start differs. The activation-gate row is the configuration of Table~\ref{tab:main}.
Avg.\ excludes clean episodes.}
\label{tab:components_detail}
\small
\setlength{\tabcolsep}{4pt}
\begin{tabularx}{\textwidth}{@{}lXrrrrrrr@{}}
\toprule
Backbone & Adaptation starts & Clean & Motion & Gaussian & Zoom & Glass & Fog & Avg. \\
\midrule
SmolVLA & Never (Frozen) & 71.5 & 63.0 & 25.0 & 29.0 & 56.5 & 46.0 & 43.9 \\
\cmidrule(l){2-9}
 & At activation gate & 70.5 & 68.5 & 43.0 & 42.0 & 58.5 & 54.0 & 53.2 \\
 & At every replan & 69.5 & 67.5 & 42.0 & 45.5 & 61.0 & 56.5 & 54.5 \\
\bottomrule
\end{tabularx}
\end{table*}

Over the evaluation of Table~\ref{tab:main}, the gate updates on 12.7\% of
nominal and 59.3\% of corrupted replans, against roughly 98\% in both conditions for always-on adaptation. Table~\ref{tab:latency_methods} reports
the resulting average per-replan cost, measured identically for every method.
Self-training shares \systems{}'s gate, so its profile matches
\systems{}'s. ActMAD updates on nearly every replan by construction, and BID
pays its candidate sampling at every replan with no gradient step.

\begin{table}[!htbp]
\centering
\caption{\textbf{Average per-replan latency across methods} (multiples of the
frozen policy; update frequencies from the evaluation of
Table~\ref{tab:main}).}
\label{tab:latency_methods}
\small
\setlength{\tabcolsep}{5pt}
\begin{tabular}{@{}lrrrr@{}}
\toprule
 & \multicolumn{2}{c}{SmolVLA} & \multicolumn{2}{c}{GR00T N1.7} \\
\cmidrule(lr){2-3}\cmidrule(lr){4-5}
Method & Nominal & Corrupted & Nominal & Corrupted \\
\midrule
Frozen & 1.00 & 1.00 & 1.00 & 1.00 \\
BID & 33.7 & 33.7 & 24.3 & 24.3 \\
ActMAD & 2.7 & 2.7 & 3.1 & 3.1 \\
Self-training & 1.4 & 2.8 & 1.5 & 3.2 \\
\systems{} (gate) & 1.5 & 2.9 & 1.4 & 3.4 \\
\systems{} (every replan) & 4.2 & 4.2 & 5.0 & 5.0 \\
\bottomrule
\end{tabular}
\end{table}

\subsection{Runtime Measurement}
\label{app:runtime}

We measure per-replan cost inside the LIBERO rollout loop on one NVIDIA H200,
one process per GPU, with CUDA synchronization at every section boundary; the
first replan of each episode is excluded as warm-up. The singular value
decomposition is computed only once when the gates are attached. A replan
without an update pays the frozen forward plus the gate's feature-change
score, which adds 3--9~ms including the embedding export
(Table~\ref{tab:latency}). An update replan adds the five-step optimization
and the regeneration of the executed chunk.

\begin{table}[!htbp]
\centering
\caption{\textbf{Per-replan latency breakdown.} Median milliseconds on one
H200, measured in the rollout loop. Optimization uses five gradient steps and
eight noise--time draws; Quiet is a replan without an update, gate included.}
\label{tab:latency}
\small
\setlength{\tabcolsep}{4.5pt}
\begin{tabular}{@{}lrrrrr@{}}
\toprule
Model & Frozen & Quiet & Optimization & Regeneration & Update replan \\
\midrule
SmolVLA & 134.8 & 143.9 & 269.3 & 143.4 & 559.3 \\
GR00T N1.7 & 139.8 & 142.8 & 414.7 & 154.4 & 723.7 \\
\bottomrule
\end{tabular}
\end{table}

\section{Supervision Diagnostics}
\label{app:supervision_diagnostics}

This section details the analyses of Sections~\ref{sec:anchoring}
and~\ref{sec:propagation}.

\subsection{Target, Alignment, and Length Controls}
\label{app:alignment_controls}

The alternative targets in Figure~\ref{fig:analysis_target_comparison} replace only the
supervised action sequence. \textbf{Another episode's plan} is the chunk
planned one replan earlier in a nominal rollout from another initial state of
the same task, so it is aligned in time but belongs to a different state.
\textbf{Time-shuffled previous chunk} reorders the previous chunk in time.
\textbf{Clean-input prediction} is the policy's prediction from the
uncorrupted version of the current observation, which is unavailable at
deployment. All rows share the adaptation gate, the update schedule, and the
random draws of the update, so they differ only in the supervised sequence;
the Frozen and leftover rows reproduce Table~\ref{tab:main} episode by
episode. Against the leftover, the clean-input prediction wins 114 episodes
and loses 77 ($p=0.009$), another episode's plan wins 62 and loses 162, and
the time-shuffled chunk wins 47 and loses 168 (both $p<10^{-10}$). Another
episode's plan does not differ from Frozen ($p=0.67$), and the time-shuffled
chunk falls slightly below it ($p=0.035$).

\begin{figure}[ht]
    \centering
    \includegraphics[width=0.32\textwidth]{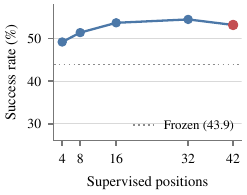}
    \caption{\textbf{Supervision length on SmolVLA.} Success when only the
    first $n$ of the 42 aligned leftover actions are supervised (the 1,000
    episodes of Table~\ref{tab:main} over five persistent corruptions).
    Supervising 16 to 42 positions performs similarly; shorter targets are
    less effective.}
    \label{fig:analysis_supervision_length}
\end{figure}

For supervision length
(Figure~\ref{fig:analysis_supervision_length}), only the first $n$ of the 42
aligned leftover actions are supervised and the rest are masked out. The 16-
to 42-position variants differ by at most 1.3 points ($p=0.73$ and $p=0.20$
for 16 and 32 against 42).

\subsection{Clean-Plan Gap under Gaussian Blur}
\label{app:mechanism_diagnostics}

\begin{figure}[ht]
    \centering
    \includegraphics[width=0.40\linewidth]{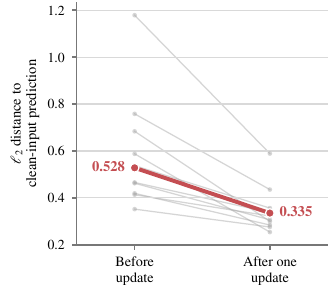}
    \caption{\textbf{One update narrows the clean-plan gap on GR00T N1.7.}
    Gaussian blur: distance to the clean-input prediction at the same
    shift-boundary state and flow-noise seed before and after one update
    ($n=200$, eight aligned positions). Thin lines are task medians; the thick
    line is the pooled median.}
    \label{fig:groot_gap_closure}
\end{figure}

Figure~\ref{fig:groot_gap_closure} repeats the SmolVLA measurement of
Figure~\ref{fig:leftover_gap_closure} on GR00T N1.7 under Gaussian blur.

\section{Activation and Gate Timing}
\label{app:ablations}
\label{app:fig1_diagnostic}

\begin{figure}[ht]
\centering
\includegraphics[width=\linewidth]{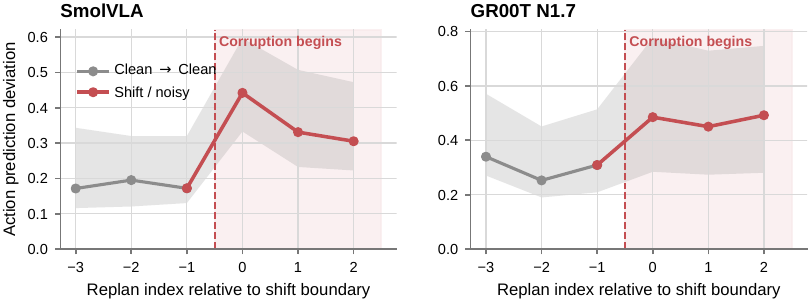}
\caption{\textbf{Action-consistency gap around the shift boundary.} Median
(line) and interquartile range (band) of the normalized RMSE between the
previous chunk's leftover and the current chunk's aligned prefix, over 150
shift boundaries per backbone. Gray marks clean-to-clean transitions and red
marks transitions at or after $t^*$. The SmolVLA panel is the right panel of
Figure~\ref{fig:lava_problem}.}
\label{fig:fig1_diagnostic_models}
\end{figure}

\paragraph{Figure~\ref{fig:lava_problem} action-consistency diagnostic.}
 The right panel of Figure~\ref{fig:lava_problem} is the SmolVLA panel of
Figure~\ref{fig:fig1_diagnostic_models}, which repeats the measurement for
GR00T N1.7. At each replan, the diagnostic measures the RMSE
between the previous chunk's unexecuted tail and the current chunk's
temporally aligned prefix after eight actions have been executed. We replay
clean reference trajectories and evaluate corrupted observations at the same
simulator states with matched flow-noise draws, without executing the
counterfactual corrupted chunks. The six arm-action dimensions are normalized
by task-specific scales estimated from calibration episodes; the gripper is
excluded. Across ten LIBERO-10 tasks, three held-out episodes per task, and
five severity-5 corruptions (150 shift boundaries events per backbone), the median
gap rises from 0.172 immediately before $t^*$ to 0.442 at $t^*$ on SmolVLA,
and from 0.309 to 0.485 on GR00T N1.7. In both cases the gap stays elevated
after $t^*$, while the preceding clean-to-clean transitions remain within
the nominal band. This action-space diagnostic is distinct
from the feature-change score used by the deployment gate.

\subsection{Gate Timing}
\label{app:detector_results}

Table~\ref{tab:trigger} summarizes when the gate activates in the evaluation
of Table~\ref{tab:main}. Clean alarm counts clean episodes with any
activation; Early counts corrupted episodes whose first activation precedes
$t^*$; each corruption column counts episodes first activated at or after
$t^*$. The remaining 1 SmolVLA and 246 GR00T episodes are never
activated: GR00T's gate misses most motion-blur and fog episodes, although it
activates on more than 90\% of the other three families. Among activations at or
after $t^*$, 98.9\% (SmolVLA) and 91.7\% (GR00T) occur at the replan of $t^*$
itself.

\begin{table*}[t]
\centering
\caption{\textbf{Gate timing on LIBERO-10} (\%). Overall divides activations
at or after $t^*$ by all 1,000 corrupted episodes; delay is the mean number of
replans between $t^*$ and activation.}
\label{tab:trigger}
\small
\setlength{\tabcolsep}{4pt}
\resizebox{\textwidth}{!}{%
\begin{tabular}{@{}lrrrrrrrrr@{}}
\toprule
Backbone & \shortstack{Clean\\alarm} & Early & Motion & Gaussian & Zoom & Glass & Fog & Overall & \shortstack{Mean\\delay} \\
\midrule
SmolVLA & 26.5 & 9.0 & 91.0 & 91.0 & 91.0 & 91.0 & 90.5 & 90.9 & 0.13 \\
GR00T N1.7 & 12.5 & 8.0 & 31.5 & 92.0 & 90.5 & 92.0 & 31.0 & 67.4 & 0.74 \\
\bottomrule
\end{tabular}
}
\end{table*}

\section{Disruption Schedules}
\label{app:scope_details}

\paragraph{Shift timing.}
Figure~\ref{fig:analysis_onset_sensitivity} fixes the shift boundary at 10\%,
25\%, 50\%, or 75\% of the clean rollout instead of sampling it, with 1,000
episodes per setting. Table~\ref{tab:shift_timing} lists the values. An
earlier boundary leaves more active replans for adaptation, and the gain
grows with them.

\begin{table}[!htbp]
\centering
\caption{\textbf{Success by shift-boundary position} (SmolVLA, \%, 1,000
episodes per row, 95\% confidence intervals in brackets). Active replans is
the mean number of replans per episode at which \systems{} updates.}
\label{tab:shift_timing}
\small
\setlength{\tabcolsep}{5pt}
\begin{tabular}{@{}lrrrr@{}}
\toprule
Shift boundary & Frozen & \systems{} & Gain & Active replans \\
\midrule
10\% & 33.5 [30.2, 36.9] & 50.0 [45.7, 54.3] & +16.5 [13.1, 19.9] & 44.3 \\
25\% & 40.0 [36.0, 44.1] & 54.1 [49.2, 59.0] & +14.1 [10.8, 17.4] & 37.5 \\
50\% & 45.1 [40.3, 49.9] & 57.0 [51.9, 62.3] & +11.9 [8.4, 15.4] & 28.4 \\
75\% & 53.9 [48.2, 59.5] & 62.5 [56.6, 68.5] & +8.6 [5.6, 11.7] & 17.3 \\
\bottomrule
\end{tabular}
\end{table}

\paragraph{Gradual and temporary corruption.}
Table~\ref{tab:schedule_detail} extends the persistent-corruption evaluation
of Table~\ref{tab:main} to two additional schedules. Gradual corruption
blends the corrupted and nominal images linearly, from the nominal image at
the first replan to full strength after eight replans, and then persists. Gradual corruption has no clear shift boundary for the gate to read.
\systems{} adapting at every replan raises success from 37.6\% to 46.0\%,
since the early leftovers come from nearly nominal frames. Temporary
corruption starts at the sampled shift boundary and clears after eight
replans. A corruption that clears also calls for a release the gate does
not provide, so both schedules are evaluated with updates at every replan
once a previous chunk is available. \systems{} continues updating after the clean input returns and
raises success from 57.2\% to 64.9\%.

\begin{table}[!htbp]
\centering
\caption{\textbf{Temporary and gradual corruption by family} (SmolVLA,
success \%, 200 episodes per corruption and method). \systems{} updates at
every replan once a previous chunk is available. Avg.\ averages the five
corrupted conditions.}
\label{tab:schedule_detail}
\small
\setlength{\tabcolsep}{4pt}
\begin{tabular*}{\linewidth}{@{\extracolsep{\fill}}llrrrrrr@{}}
\toprule
Schedule & Method & Motion & Gaussian & Zoom & Glass & Fog & Avg. \\
\midrule
\multirow{2}{*}{Temporary} & Frozen & 63.5 & 48.5 & 51.5 & 64.5 & 58.0 & 57.2 \\
 & \systems{} & 68.0 & 59.5 & 62.0 & 69.0 & 66.0 & 64.9 \\
\midrule
\multirow{2}{*}{Gradual} & Frozen & 62.5 & 13.5 & 21.0 & 50.5 & 40.5 & 37.6 \\
 & \systems{} & 64.5 & 28.0 & 35.5 & 59.0 & 43.0 & 46.0 \\
\bottomrule
\end{tabular*}
\end{table}

\paragraph{Corruption from the first replan.}
Table~\ref{tab:from_start} applies each corruption from replan 0, so no
leftover is planned from a nominal frame. Adapting at every replan once a
previous chunk exists leaves overall success nearly unchanged (32.5\% versus
32.9\%).

\begin{table}[!htbp]
\centering
\caption{\textbf{Corruption from the first replan} (SmolVLA, success \%, 200
episodes per family). \systems{} updates at every replan once a previous
chunk is available. Avg.\ averages the five corrupted conditions.}
\label{tab:from_start}
\small
\setlength{\tabcolsep}{4pt}
\begin{tabular*}{\linewidth}{@{\extracolsep{\fill}}lrrrrrr@{}}
\toprule
Method & Motion & Gaussian & Zoom & Glass & Fog & Avg. \\
\midrule
Frozen & 58.0 & 3.5 & 11.0 & 51.0 & 39.0 & 32.5 \\
\systems{} & 56.5 & 8.0 & 14.0 & 51.0 & 35.0 & 32.9 \\
\bottomrule
\end{tabular*}
\end{table}

\section{Limitations and Future Work}
\label{app:limitations}

\systems{}'s supervision comes from the leftover, so the method is only as
good as the leftover it receives. When the disruption is present from the
start of the episode, no nominal-side plan exists and the leftover offers
little to teach from. Disruptions that invalidate the plan itself, such as
moved objects or changed goals, are also beyond what leftover supervision can
correct. Separately, the adaptation gate is a deployment trigger. It reads
transitions between consecutive
replans, so it can miss shifts that produce no clear transition, for example
those that arrive gradually. Any activation rule can drive the same
leftover-supervised update. Improving activation is therefore left to future
work, for example by accumulating evidence across replans. Finally,
adaptation is learned at test time, and an update replan pays for gradient
steps and chunk regeneration. Reducing this cost with fewer optimizer steps
or lighter update rules is also left to future work.

\end{document}